\documentclass{article}
\usepackage{spconf,amsmath,graphicx,bbm}
\usepackage{times}
\usepackage{epsfig}
\usepackage{amssymb,color}
\usepackage{enumitem} 
\setitemize{noitemsep,topsep=0pt,parsep=0pt,partopsep=0pt}
\usepackage{subcaption}

\title{\normalsize Joint learning of assignment and representation for biometric group membership}
\name{\begin{tabular}{c}Marzieh Gheisari, Teddy Furon, Laurent Amsaleg, \thanks{Research supported by the ERA-Net project ID\_IoT 20CH21\_167534.} \end{tabular}}
\address{Univ Rennes, Inria, CNRS, IRISA, France\\
	{\small \{ \texttt{marzieh.gheisari-khorasgani, teddy.furon}\}@\texttt{inria.fr}},
	{\small \texttt{laurent.amsaleg}@\texttt{irisa.fr}}\\
	}
\begin{document}
\newcommand\vect[1]{\mathbf{#1}} 
\newcommand\func[1]{\mathsf{#1}} 
\def \group {\mathcal{S}}
\def \x {\vect{x}} 
\def \y {\vect{q}} 
\def \z {\vect{z}} 
\def \f {\vect{f}}
\def \X {\vect{X}}
\def \Y {\vect{Y}}
\def \Z {\vect{Z}}
\def \w {\vect{u}} 
\def \m {\vect{m}}
\def \e {\vect{e}}
\def \h {\vect{e}}  
\def \a {\vect{a}}
\def \r {\vect{r}}
\def \n {\vect{n}} 
\def \W {\vect{W}} 
\def \G {\vect{G}} 
\def \real {\mathbb{R}} 
\def \dim {d} 
\def \rep {\vect{r}} 
\def \hyp {\mathcal{H}} 
\def \one {\vect{1}}
\def \W {\vect{W}}
\def \H {\vect{E}}
\def \R {\vect{R}}
\def \Av {\vect{A}}
\def \A {\mathcal{A}}
\def \Pr {\mathbb{P}} 
\def \Pfn {P_{\mathsf{fn}}}
\def \Pfp {P_{\mathsf{fp}}}
\def \Ptp {P_{\mathsf{tp}}}
\def \pfn {p_{\mathsf{fn}}}
\def \ptp {p_{\mathsf{tp}}}
\def \pfp {p_{\mathsf{fp}}}
\def\un{{\mathbbm{1}}}
\def\E{{\mathbbm{E}}}
\def\Var{{\mathbbm{V}}}
\def\0{{\vect{0}}}
\def\etal{{\it et al.}}
\def\ie{{\it i.e.}}
\def\AoE{\text{\tiny{AoE}}}
\def\EoA{\text{\tiny{EoA}}}
\def\yg{\mathcal{Y}_g}
\def \Q {\vect{Q}}
\newcommand\alert[1]{{\color{red}{#1}}}

\newcommand\blfootnote[1]{%
	\begingroup
	\renewcommand\thefootnote{}\footnote{#1}%
	\addtocounter{footnote}{-1}%
	\endgroup
}

\maketitle
\begin{abstract}
This paper proposes a framework for group membership protocols preventing the curious but honest server from reconstructing the enrolled biometric signatures and inferring the identity of querying clients. This framework learns the embedding parameters, group representations and assignments simultaneously. Experiments show the trade-off between security/privacy and verification/identification performances.
\end{abstract}
\vspace{-10pt}
\section{Introduction}
\label{sec:Introduction}
\vspace{-7pt}
Group membership verification is a procedure checking
whether an item or an individual is a member of a group. If
membership is positively established, then an access to some
resources (buildings, wifi, payment, conveyor units, \ldots) is granted; otherwise the access is refused.
Being granted with this shared privileged access requires that the members of the group could be distinguished from non-members, but it does not require to distinguish members from one another. Indeed, privacy concerns suggest that the verification should be carried-out anonymously.\\

\vspace{-10pt}
This paper studies group verification and also group identification. In this later setup, there are several groups of members and one needs to identify in which group the user is belonging to. 
This paper focuses on privacy preserving group identification procedure where group identity of a member is found without disclosing the identity of that individual.\\


\vspace{-12pt}
In computer vision, it is very common to aggregate signals into one representation ~\cite{Sivic:2003qp,jegou:inria-00633013,Perronnin:2007qm}, but they do not consider security or privacy.
For instance, in~\cite{iscen:hal-01481220}, Iscen \etal\ use the \emph{group testing} paradigm to pack a random set of image signatures into a unique high-dimensional vector where the similarities between the original non-aggregated signatures and a query signature is preserved through the aggregation.\\

\vspace{-12pt}
Recently ~\cite{Gheisari2019icassp, Gheisari_2019_CVPR_Workshops,Gheisari2019WIFS} proposed a framework based on aggregation and embedding of several biometric signatures into a unique vector representing the members of a group. It has been demonstrated that this allows a good assessment of the membership property at test time provided that the groups are small. It has also been shown that this provides privacy and security.  Privacy is enforced because it is impossible to infer from the aggregated feature which original signature matches the one used to probe the system. Security is preserved since nothing meaningful leaks from embedded data~\cite{Razeghi2017wifs,Razeghi2018icassp}.\\

\vspace{-12pt}
This paper revisits the core mechanism proposed by \cite{Gheisari_2019_CVPR_Workshops}. 
That work, however, is deterministic in the sense that it learns group representations based on predefined groups.
This paper shows that learning jointly the group representations and group assignments results in better performance without damaging the security. This adresses scenarios where the number of members is too big. Their signatures can not be packed into one unique group representation with a technique like~\cite{Gheisari_2019_CVPR_Workshops}.
Therefore, members are automatically assigned to different groups.
A light cryptographic protocol is deployed to secure their privacy during group verification.\\

\vspace{-6pt}
\vspace{-10pt}
\section{Group membership}
\label{sec:ProblemFormulation}
\vspace{-10pt}

\def\p {\mathbf{p}}
\subsection{Notations}
The embedding, the assignment, and the group representations are learned jointly at enrolment, and given to a server.
Biometric signatures are modelled as vectors in $\real^{\dim}$. $\X \in \real^{d\times N}$ is the matrix of the signatures to be enrolled into $M$ groups. The group representations are stored column wise in $\ell\times M$ matrix $\R$.
The group representations are quantized and sparse \ie, $\r_g\in \A^{\ell}$ with $\A=\{-1,0,+1\}$ and $\|\r_{g}\|_{0}\leq S<\ell$, $\forall g\in[M]$.\\

\vspace{-12pt}

At query time, the user computes a sparse representation of his biometric signature $\y \in \real^\dim$.
For that purpose, function $\func{e}:\real^{\dim}\to \mathcal{A}^{\ell}$ maps a vector to a sequence of $\ell$ discrete symbols.  We use the sparsifying transform coding~\cite{Razeghi2017wifs, Razeghi2018icassp}:
$\p:=\func{e}(\y) = \func{T}_S(\mathbf{W}^\top\y)$.
After projecting $\y \in \real^{d}$ on the column vectors of $\mathbf{W} \in \real^{\dim \times \ell}$,  the output alphabet $\A$ is imposed by the ternarization function $\func{T}_S$:  The $\ell-S$ components having the lowest amplitude are set to 0.  The $S$ remaining ones are quantized to +1 or -1 according to their sign.

\vspace{-10pt}
\subsection{Formulation of the optimization problem} 
\label{sec:ProposedMethod}
\vspace{-8pt}
Our group membership protocol aims at jointly learning the partition, the embedding and the group representations.
The key is to introduce the auxiliary data $\H=[\h_1$, \ldots, $\h_N] \in \A^{\ell\times N}$ the hash codes of enrolled signatures and $\Y\in \real^{M\times N}$ the group indicator matrix ($y_{i,j}=1$ if $\h_j$ is assigned to $i$-th group). Then, the optimization problem is composed of a cost for embedding $C^{E}$ and a cost for partitioning $C^{A,G}$:
\vspace{-6pt}
\begin{equation}
\min_{\W,\R,\Y} C^{E}(\X,\W,\H)+ C^{A,G}(\H,\Y,\R) ,
\label{eq:OptimizationProblemEoA}
\vspace{-5pt}
\end{equation}
\vspace{-6pt}
The embedding cost is the loss for quantizing signatures:
\vspace{-3pt}
\begin{eqnarray}
\label{eq:QuantizationLoss}
C^{E}(\X,\W,\H)& :=&  \sum_{i=1}^{N} \left\Vert\h_i -\W^\top \x_i \right\Vert^2_2.
\end{eqnarray}
\vspace{-1pt}

The assignment aims at grouping together signatures sharing similar hash codes: the overall dissimilarity between members and their group representation is minimized while the separation between two groups is maximized. 
%
Inspired by Linear Discriminant Analysis, we consider variance to measure dissimilarity.
The within group scatter matrix $\vect{S}_w$ and the between group scatter matrix $\vect{S}_b$ are defined as
\begin{eqnarray*}
\vect{S}_w&=&\sum_{g=1}^{M}\sum_{i\in \yg}(\h_{i}-\rep_g)(\h_{i}-\rep_g)^\top=(\H-\R\Y)(\H-\R\Y)^\top\\
\vect{S}_b&=&\sum_{g=1}^{M} \rep_g \rep_g^\top=\R\Y(\R\Y)^\top
\end{eqnarray*}
where $\yg=\{i\in[N]:y_{g,i}=1\}$.
The cost for partitioning is $C^{A,G} = \lambda Tr(\vect{S}_w)-\gamma Tr(\vect{S}_b)$ for some $\lambda$, $\gamma$ in $\real_+$.


In the end, the objective function is formulated as:
\vspace{-8pt}
\begin{equation}
\label{eq:LDA_obj}
\begin{aligned}
& \underset{\W,\R,\Y} {\text{min}}
& &  \|\H - \W^{\top}\X\|_{F}^{2}+\lambda Tr(\vect{S}_w)-\gamma Tr(\vect{S}_b)\\
&\text{s.t.}
& & \W^T\W=\vect{I}_\ell\\
& & & \Y \in \{0,1\}^{M \times N}, \;\; 
\left\Vert\vect{y}_i \right\Vert_1 =1 \; \forall i \in [N]\\
& & & \h_i \in \A^{\ell}, \; \; \left\Vert \h_i \right\Vert_0\leq S\\
& & & \r_g \in \A^{\ell}, \; \; \left\Vert \rep_g \right\Vert_0\leq S
\end{aligned}
\end{equation}
\vspace{-2pt}
The constraint on $\Y$ ensures that each signature belongs to exactly one group.

\subsection{Suboptimal solution}
The solution of \eqref{eq:LDA_obj} is found by iterating the following steps:

\textbf{$\W$-Step.} We fix $\H$, $\R$, $\Y$ and update $\W$ by solving:
\vspace{-8pt}
\begin{equation}
\label{eq:OptW}
\begin{aligned}
& \underset{\W} {\min}
& &  \left\Vert\H -\W^\top \X \right\Vert^2_F\\
&\text{s.t.}
& & \W^\top\W=\vect{I}_\ell
\end{aligned}
\end{equation}
This problem is a least square Procruste problem with orthogonality constraint.
By setting $\vect{S}:=\X\H^\top$,~\cite{Schonemann1966} shows that $\W=\vect{UV}^\top$, where $\vect{U}$ contains the eigenvectors corresponding to the $\ell$  ($\ell < \dim$) largest eigenvalues of $\vect{SS}^\top$ and $\vect{V}$ contains the eigenvectors of $\vect{S}^\top\vect{S}$.

\textbf{E-Step.} Given $\W$, $\Y$ and $\R$, \eqref{eq:LDA_obj} amounts to:
\begin{equation}
\label{eq:OptW}
\begin{aligned}
& \underset{\H} {\min}
& &  \left\Vert\H -\W^\top \X \right\Vert^2_F+\lambda \left\Vert\H -\R \Y \right\Vert^2_F\\
&\text{s.t.}
& & \h_i \in \A^{\ell}, \; \; \left\Vert \h_i \right\Vert_0\leq S
\end{aligned}
\end{equation}
We first find the solution relaxing the constraints and then apply ternarization function $\func{T}_S$ 
to obtain sparse codes:
\vspace{-5pt}
\begin{equation}
\label{eq:OptE}
\begin{aligned}
& & \H=\func{T}_S(\W^\top\X+\lambda \R\Y).
\end{aligned}
\end{equation}

\textbf{(R,Y)-Step.} When fixing $\W$ and $\H$,
the assignment and group representations are found by minimizing:
\begin{equation}
\label{eq:OptRY}
\begin{aligned}
& \underset{\R,\Y} {\min}
& &  \left\Vert\H -\R \Y \right\Vert^2_F-\frac{\lambda}{\gamma} Tr (\R\Y\Y^\top\R^\top)\\
&\text{s.t.}
& & \Y \in \{0,1\}^{M \times N}, \;\; 
\left\Vert\vect{y}_i \right\Vert_1 =1 \; \forall i \in [N]\\
& & & \r_g \in \A^{\ell}, \; \; \left\Vert \rep_g \right\Vert_0\leq S
\end{aligned}
\end{equation}
As $\H$ is fixed, $Tr(\H\H^\top)$ is irrelevant to $\Y$, thus minimizing \eqref{eq:OptRY} is equivalent to:
\vspace{-10pt}
\begin{equation}
\label{eq:kmeans}
\begin{aligned}
& \underset{\vect{R},\vect{Y}} {\text{min}}
& &  \left\Vert \frac{\lambda}{\lambda-\gamma}\H-\vect{RY} \right\Vert^2_F.
\end{aligned}
\end{equation}
Relaxing the ternarization constraint, \eqref{eq:kmeans} is solved by a k-means clustering algorithm, \ie iteratively:
\begin{itemize}
	\item \textit{Update assignments}: Each item is assigned to its nearest group representative.
	\item \textit{Update centroids}: $g$-th centroid is the mean of all $\tilde{\h}_i$ in group $g$. 
\end{itemize}
Then the group representation $\r_g$ is found by applying ternarization function on $g$-th centroid.

\begin{figure*}[tb]%
	\vspace{-30pt}	
	\centering
	\includegraphics[width=0.95\linewidth]{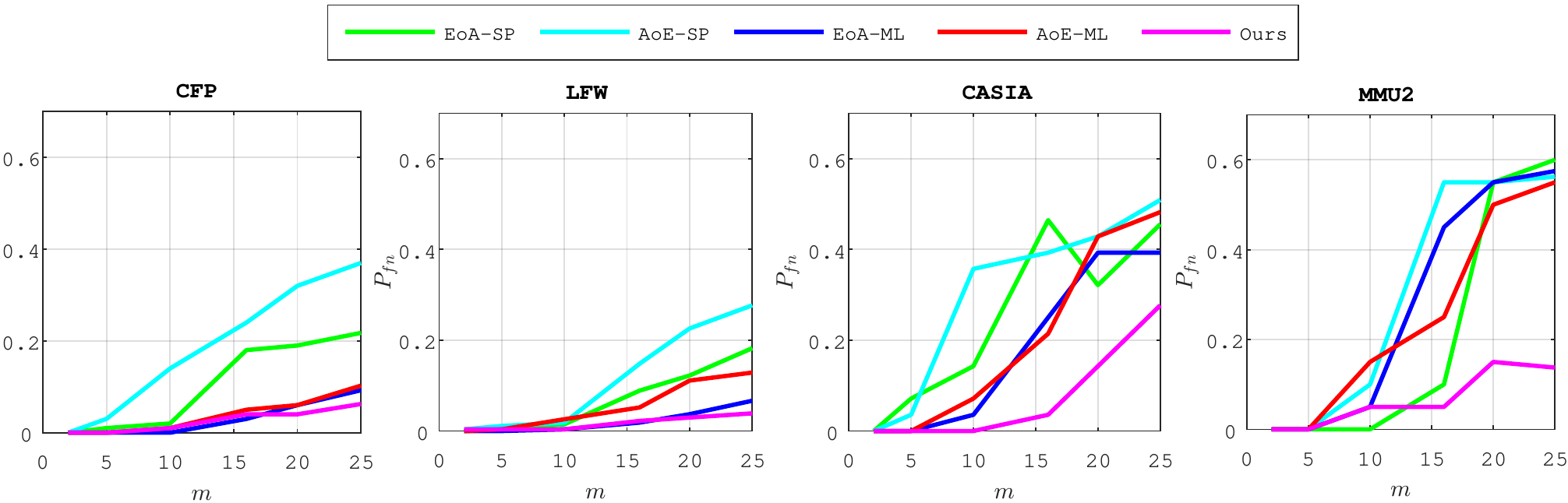}
	\vspace{-8pt}	
	\caption{Performances comparison for varying group size $m$. $\Pfn$ at $\Pfp=0.05$ for group verification.}
	\label{fig:verif}%
\end{figure*}

\vspace{-4pt}
\section{Experiments}
\label{sec:Experiments}
\vspace{-10pt}
This section presents the datasets used in our experiments and investigates the performance of the proposed method for two application scenarios. We compare our scheme with EoA-SP, AoE-SP~\cite{Gheisari2019icassp} and EoA-ML, AoE-ML ~\cite{Gheisari_2019_CVPR_Workshops}.
For the baselines $N$ individuals of each dataset are enrolled into $M$ random groups but for our scheme the algorithm learns how to partition the enrolled templates.

\vspace{-10pt}

\subsection{Datasets}
\vspace{-4pt}
\subsubsection{Face Datasets}
\vspace{-6pt}
Face descriptors are obtained from a pre-trained network based on VGG-Face architecture~\cite{parkhi2015deep} followed by PCA and then $L_{2}$-normalization with $\dim = 1,024$.\\
\vspace{-8pt}

\textbf{LFW~\cite{huang2008labeled}.}
These are pictures of celebrities in all sort of viewpoint and under an uncontrolled environment. We use pre-aligned LFW images.
The enrollment set consists of $N = 1680$ individuals with at least two images in the LFW database.
One random template of each individual is enrolled in the system, playing the role of $\x_{i}$.
Some other $N_q = 263$ individuals were randomly picked in the database to play the role of impostors.\\

\vspace{-8pt}
\textbf{CFP~\cite{sengupta2016frontal}.}
These are frontal and profile views of celebrities taken in an uncontrolled environnement.
We use $N = 400$ frontal images to be enrolled in the system.
The impostor set is a random selection of $N_q = 100$ other individuals.
\vspace{-12pt}
\subsubsection{IRIS Datasets}

\vspace{-6pt}
Iris images are prepossessed by the following steps: iris localization, iris normalization and image enhancement. 
Then the feature vectors are extracted by Gabor filters.\\

\vspace{-10pt}
\textbf{CASIA-IrisV1~\cite{Irisv1}.}
The database includes 756 iris images from 108 eyes of Chinese persons. The images stored in the database were captured within a highly
constrained capturing environment.
3 images were collected in a first session and 4 images in a second session. 
The database is created by randomly sampling $N = 80$ individuals to be enrolled, and $N_q=28$ impostors.\\

\vspace{-10pt}
\textbf{MMU2~\cite{mmu2}.}
This dataset contains $995$ images corresponding to $100$ people with different age and nationality from Asia, Middle East, Africa and Europe.
Each of them contributes to 5 iris images for each eye.
We exclude 5 left eye iris images due to cataract disease.

\vspace{-10pt}
\subsection{Group Verification}
\vspace{-5pt}
A user claims she/he belongs to group $g$.
This claim is true under hypothesis $\hyp_{1}$ and false under hypothesis $\hyp_{0}$ (\ie\ the user is an impostor).  Her/his signature $\y$ is embedded into $\p=\func{e}(\y)$, and $(\p, g)$ is sent to the system, which compares $\p$ to the group representation $\r_{g}$.  The system accepts ($t = 1$) or rejects ($t=0$) the claim. This is a two hypothesis test with two probabilities of errors: $\Pfp:=\Pr(t=1|\hyp_{0})$ is the false positive rate and $\Pfn:=\Pr(t=0|\hyp_{1})$ is the false negative rate.  The figure of merit is $\Pfn$ when $\Pfp = 0.05$.\\

\vspace{-10pt}    
Fig.~\ref{fig:verif} compares the performance of our scheme with baselines for group membership verification.
Totally our scheme gives a better verification performance especially on CASIA.
Since our method tries to simultaneously learn group representations and assignment, it aggregates similar embedded vectors and this looses less information.\\

\vspace{-10pt}    
Note that, although LFW and CFP are difficult datasets due to the "in the wild" variations, the group membership verification task is handled well even for large group sizes. This is not the case for iris datasets. As mentioned before, we make use of VGG-Face for face datasets while for iris, traditional feature extraction algorithms are used. So, the big difference in overall analysis shows how the feature space affect the performance of group membership tasks.
\vspace{-12pt}
\subsection{Group Identification}
\vspace{-6pt}    
The scenario is an open set identification where the querying user is either enrolled or an impostor.  The system proceeds in two steps. First, it decides whether or not this user is enrolled.  This is  verification as above, except that the group is unknow: The system computes $\delta_{j} = \|\p-\rep_{j}\|$, $\forall j\in[M]$, and accepts $(t=1)$ if the minimum of these $M$ distances is below a given threshold $\tau$.  The figure of merit is $\Pfn$ when $\Pfp = 0.05$.

When $t=1$, the system proceeds to the second step. The estimated group is given by $\hat{g} = \arg \min_{j\in[M]} \delta_{j}$. The figure of merit for this second step is $P_{\epsilon}:= \Pr(\hat{g}\neq g)$ or the Detection and Identification Rate $DIR := (1-P_{\epsilon})(1-\Pfn)$.\\

\begin{figure}[tb]%
	\centering
	\includegraphics[width=0.95\linewidth,height=5cm]{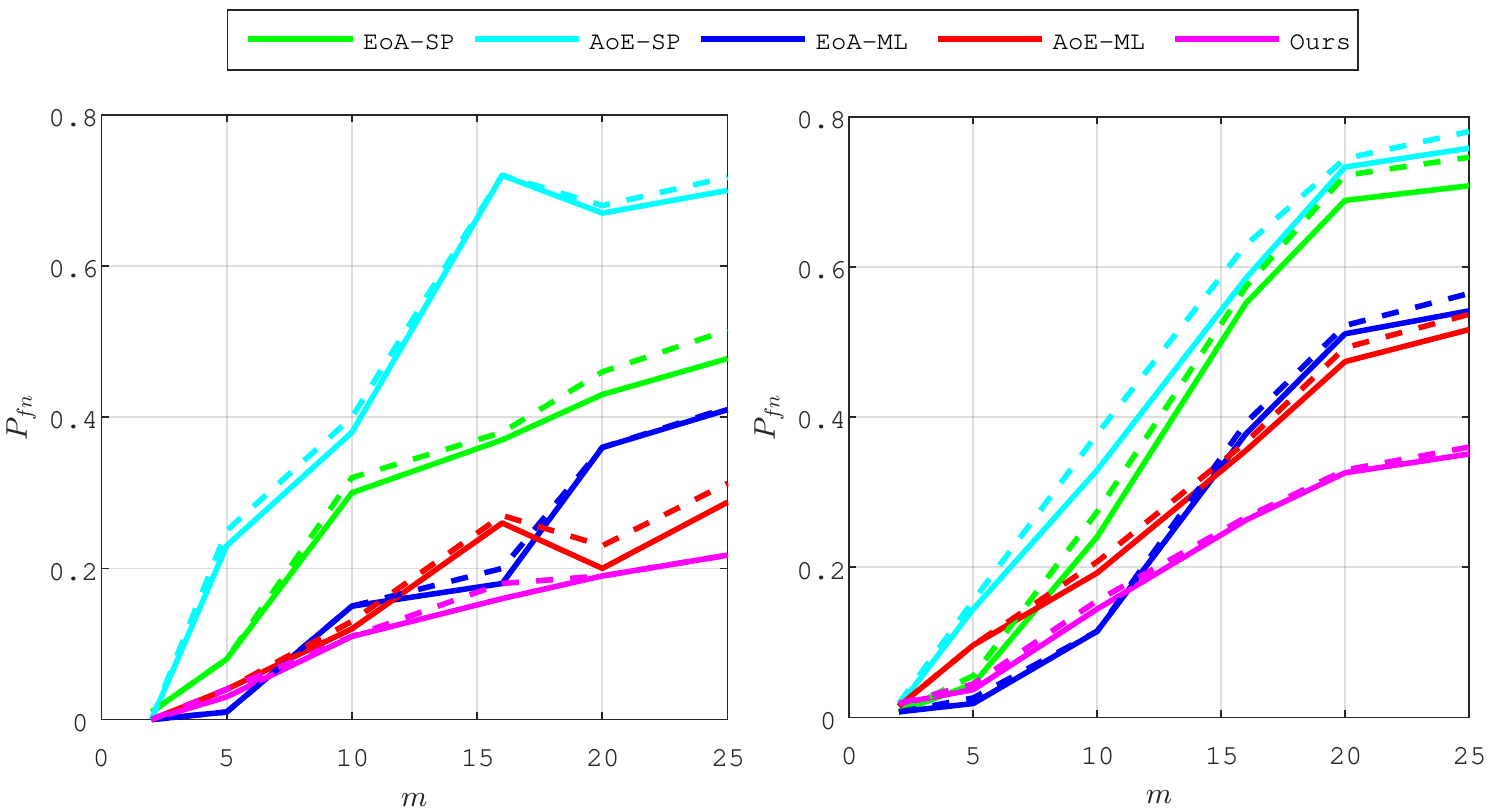}
	\caption{Performances comparison for varying group size $m$ on group identification for CFP(left) and LFW(right). $\Pfn$ at $\Pfp=0.05$ for the first step of group identification (solid) and $P_{\epsilon}$ for the second step of group identification (dashed).}
	\vspace{-13pt}
	\label{fig:iden}%
\end{figure}

\vspace{-6pt}
Fig.~\ref{fig:iden} shows that our scheme brings improvement compared to the baselines and the improvement is also better as the size of groups increases.\\

\begin{figure}[tb]%
	\centering
	\includegraphics[width=0.95\linewidth,height=5cm]{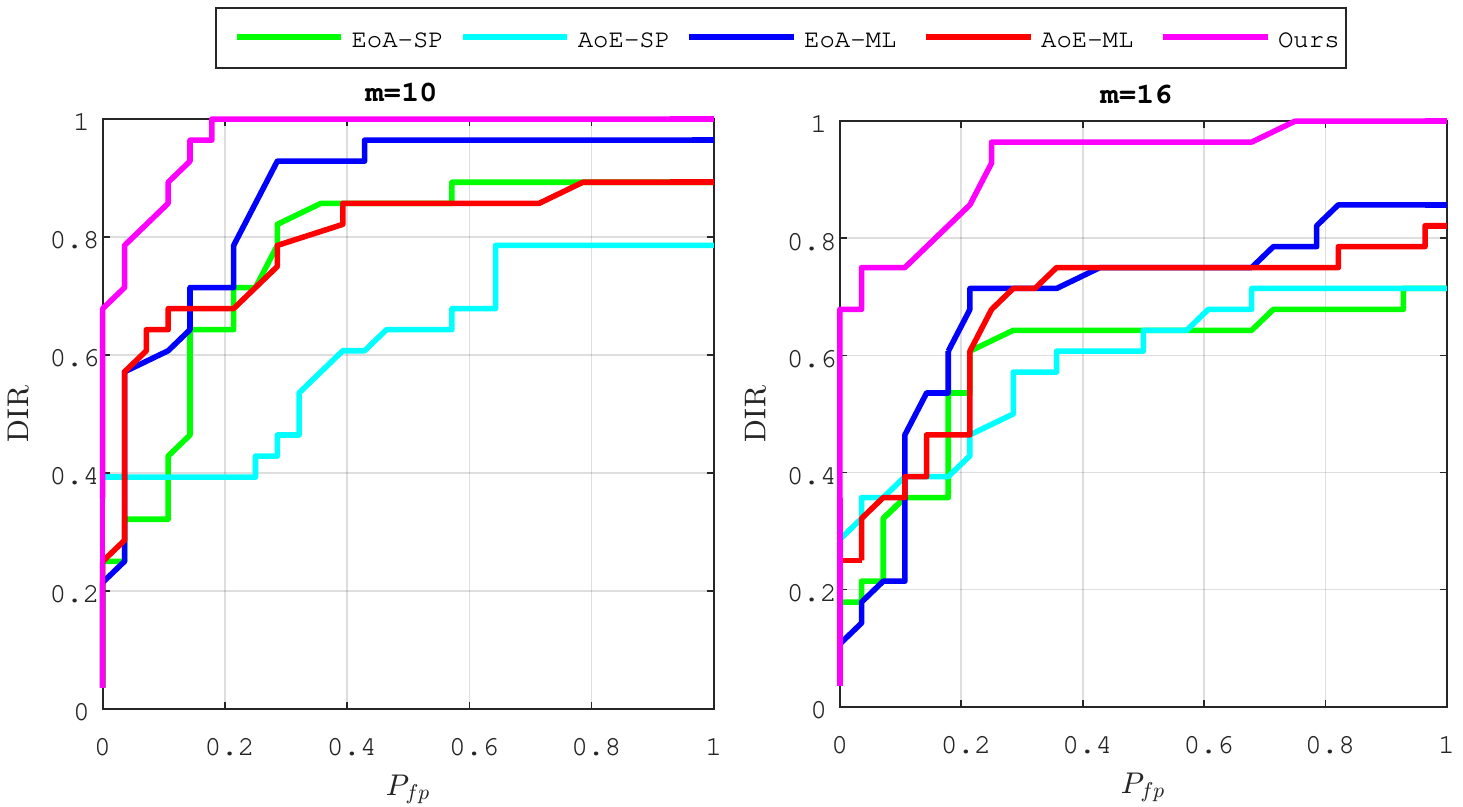}
	\caption{The Detection and Identification Rate ($DIR$) vs.\ $\Pfp$ for group identification on CASIA-IRISV1.}
	\label{fig:DIR}%
\end{figure}

\vspace{-12pt}
The impact of the group size on DIR is illustrated in Fig.~\ref{fig:DIR}. Obviously, packing more signatures into one group representation is detrimental. It gets worse when the queries are not well correlated with the enrolled signature.

\vspace{-8pt}
\subsection{Security and Privacy Analysis}
\vspace{-4pt}
A curious server can only reconstruct a single vector $\hat{\r}_g=\func{rec}(\r_g)$ from the group representation $\r_g$, and this vector serves as an estimation of any signature in the group. We measure the security by the mean square error over the dataset:
\begin{equation}
\label{eq:MSE_S}
\func{MSE}_{S} = (dN)^{-1}\sum_{g=1}^{M}\sum_{i=1}^{|\yg|}\E(\|\x_{i}  -\hat{\r}_{g}\|^{2}).
\end{equation}

\begin{figure}[tb]%
	\centering
	\includegraphics[width=0.95\linewidth,height=5cm]{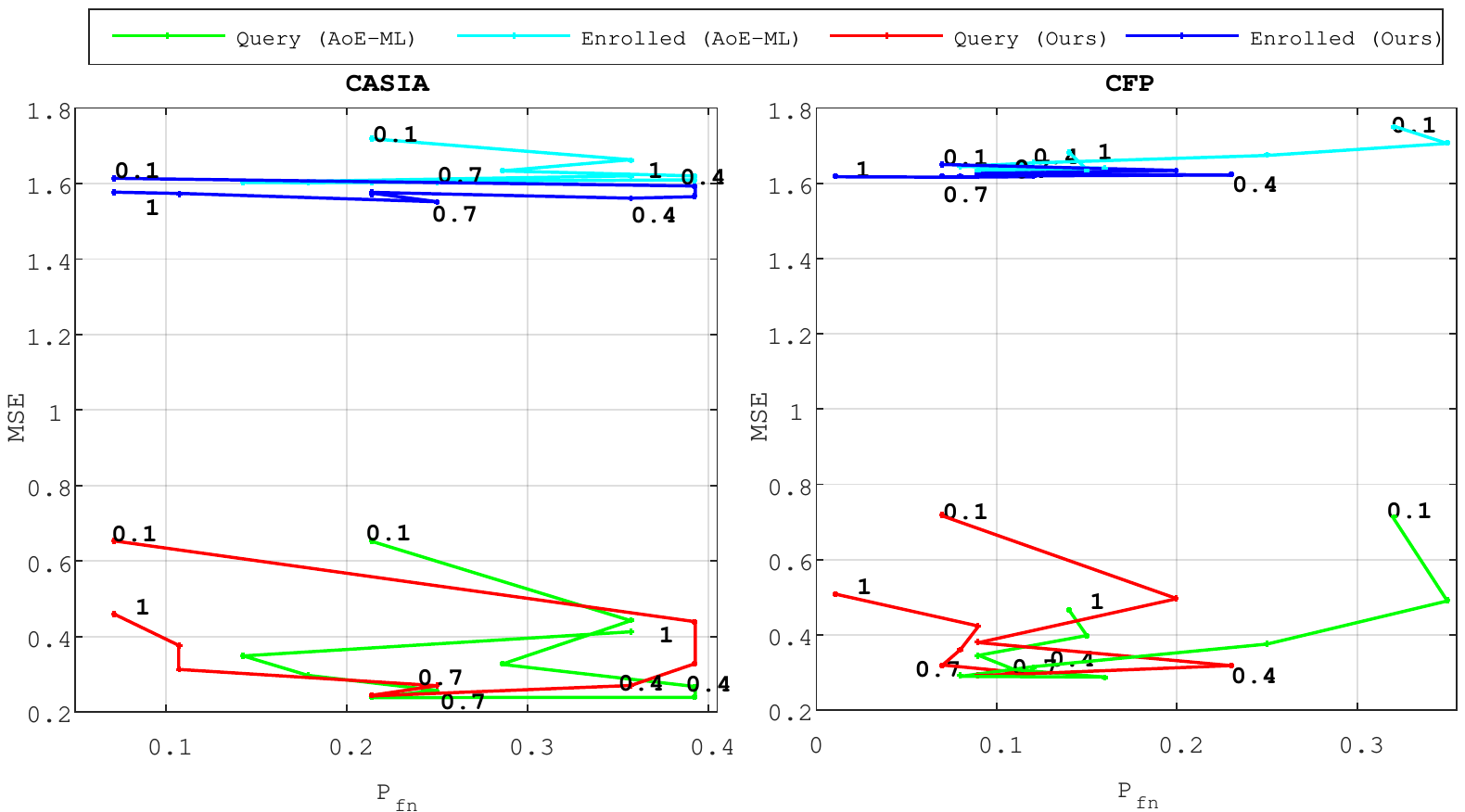}
	\caption{Investigation of trade-off between security and performance for varying sparsity level $S$ on CFP (with $m=25$) and CASIA-IrisV1 (with $m=16$).}
	\label{fig:MSE}%
	\vspace{-12pt}
\end{figure}

\vspace{-8pt}
For the of privacy of query template, a curious server can reconstruct the query template $\y$ from its embedding:
\vspace{-8pt}
\begin{equation}
\label{eq:MSE_P}
\func{MSE}_P = d^{-1}\E(\|\Q -\func{rec}(\func{e}(\Q))\|^{2}),
\end{equation}
\vspace{-2pt}
These reconstructions are possible only if matrix $\W$ is known.
This is not the case in practice, so we give here an extra advantage to the curious server. 
Figure~\ref{fig:MSE} compares security with AoE-ML~\cite{Gheisari_2019_CVPR_Workshops}
where the assignment was imposed randomly, \ie\ not learned. Different levels of sparsity are tested.
The reconstruction error of queries are close in either case, yet learning the assignment improves verification performance.
Reconstructing enrolled signatures is more difficult due to the aggregation.
However, learning the assignment by similarity correspondence in the embedded domain decreases the security slightly while improving the performance a lot. 

\section{Security Protocols}
\label{sec:SecProtocol}
\vspace{-6pt}
This section gives an example of a cryptographic protocol exploiting the group representations.
The experimental section showed that grouping secures the enrolled signatures,
but ternarization alone provides less protection to the query.
Therefore, this protocol strengthens the protection of the querying user.
For security reason, the server only manipulates query and the distances in the encrypted domain. 
For privacy reason, the server only learns that the query is close enough to one group representation,
but it cannot tell which group exactly.
We assume honest but curious user and server. 

This protocol also justifies choices of our scheme: Queries  and group representations are heavily quantized onto a small alphabet $\A$. They are long vectors but sparse: only $S$ components will be processed in the encrypted domain.
Moreover, we have $\|\p-\r\|^2\in[0,2S]$. These facts ease the use of partial homomorphic encryptions
with limited module, whence a low complexity and expansion factor.
The group representations remain in the clear on the server side, and we do not need fully homomorphic encryption.

The user generates a pair of secret and public keys $(sk_U, pk_U)$ for an additive homomorphic cryptosystem $e(\cdot)$ (say~\cite{10.1007/3-540-48910-X_16}), and sends the query encrypted 
component-wise.
The server computes its correlation with group representation $\r_g$:
\begin{equation}
e(\p^\top\r_g,pk_U) = \prod_{i:r_g(i)\neq0} e(p(i),pk_U)^{r_g(i)}.
\end{equation}

The server also generates a key pair $(sk_S,pk_S)$ for a multiplicative homomorphic 
cryptosystem $E(\cdot)$ (say \cite{1057074}), and sends the user $(E(e(\p^\top\r_g,pk_U),pk_S))_g$.
The user randomly permutes the order of these quantities and masks them by multiplying them by $E(1,pk_S)$.
This yields another semantically secure version of the ciphertexts  thanks to the multiplicative homomorphy of $E(\cdot)$.
The server decrypts $(e(\p^\top\r_{k},pk_U))_k$, but the permutation prevents connecting $k$ back to the group index $g$.
Again thanks to homomorphy, the server computes $(e(a_k(2S - 2\p^\top\r_{k} - \tau)+b_k),pk_U))_g$ where $(a_k,b_k)$ are random signed integers. The user decrypts and sends $(a_k(\|\p-\r_k\|^2 - \tau)+b_k)_k$ to the server.
The user cannot guess the distances $\|\p-\r_k\|^2$ thanks to the masking $(a_k,b_k)_k$, not even the sign of $(\|\p-\r_k\|^2 - \tau)$. The server can do this (since it knows $(a_k,b_k)$) and thus learns whether there is one group where $(\|\p-\r_k\|^2 - \tau)$ is negative.
\vspace{-6pt}

\section{Conclusion}
We proposed a method for group membership verification and identification jointly learning group representations and assignment.
The idea is to minimize the overall distance between group members while maximizing the separation between groups in the embedded domain. Yet, the method still has some rigidness: the prototyping of the embedding (the sparse ternary quantization), considering mean as group centroids, and assigning a signature to only one group.
\newpage
\bibliographystyle{IEEEbib}
\bibliography{ms}

\end{document}